\pdfoutput=1

\documentclass[11pt]{article}

\usepackage[]{acl}

\usepackage{times}
\usepackage{latexsym}
\usepackage{xspace}
\usepackage{booktabs}
\usepackage{amsmath}
\usepackage{MnSymbol}
\usepackage{graphicx}
\usepackage{multirow}
\usepackage{textcomp}

\usepackage[T1]{fontenc}

\usepackage[utf8]{inputenc}

\usepackage{microtype}

\newcommand{\balance}{\textsc{Balance}\xspace}
\newcommand{\resistance}{\textsc{Resistance}\xspace}

\newcommand{\verticality}{\textsc{Verticality}\xspace}

\newcommand{\centerperiphery}{\textsc{Center--Periphery}\xspace}

\newcommand{\linkage}{\textsc{Linkage}\xspace}

\newcommand{\support}{\textsc{Support}\xspace} 
\newcommand{\pressure}{\textsc{Pressure}\xspace}
\newcommand{\constraint}{\textsc{Constraint}\xspace}
\newcommand{\covering}{\textsc{Covering}\xspace}
\newcommand{\visibility}{\textsc{Visibility}\xspace}

\title{Exploring Spatial Schema Intuitions in Large Language and Vision Models}

\author{Philipp Wicke\thanks{\hspace{.4em} Both authors contributed equally.}* \\
  Ludwig-Maximilian-University, Munich\\
  Munich Center for Machine Learning (MCML) \\
  \texttt{pwicke@cis.lmu.de} \\\And
  Lennart Wachowiak* \\
  King's College London \\
  Imperial College London \\
 \texttt{lennart.wachowiak@kcl.ac.uk} \\}

\begin{document}
\maketitle
\begin{abstract}
Despite the ubiquity of large language models (LLMs) in AI research, the question of embodiment in LLMs remains underexplored, distinguishing them from embodied systems in robotics where sensory perception directly informs physical action.
Our investigation navigates the intriguing terrain of whether LLMs, despite their non-embodied nature, effectively capture implicit human intuitions about fundamental, spatial building blocks of language. We employ insights from spatial cognitive foundations developed through early sensorimotor experiences, guiding our exploration through the reproduction of three psycholinguistic experiments. Surprisingly, correlations between model outputs and human responses emerge, revealing adaptability without a tangible connection to embodied experiences. Notable distinctions include polarized language model responses and reduced correlations in vision language models. This research contributes to a nuanced understanding of the interplay between language, spatial experiences, and the computations made by large language models.\footnote{Project site: \url{https://cisnlp.github.io/Spatial_Schemas/}}%
\end{abstract}

\section{Introduction}



Large language models (LLMs) excel in varied NLP tasks like text generation, sentiment analysis, or summarization. Nonetheless, an underexplored facet in the study of LLMs pertains to the concept of embodiment. Unlike embodied systems in robotics, where the physical form plays a central role in shaping the system's abilities, LLMs lack a direct connection between sensory perception and physical action. Within this context, we investigate the extent to which LLMs, despite their lack of direct embodiment, might capture the implicit, often sensory-derived, conceptual structures that underlie human language and cognition. In our analysis, we make use of image schema theory, which provides a set of spatio-temporal cognitive building blocks that are learned in early infancy based on reoccurring sensorimotor experiences \cite{lakoff2008metaphors,barsalou2008grounded,johnson2013body}. For example, the image schema \support is learned by observing and experiencing objects like tables or chairs supporting other objects. According to theory, the same image schema is reused to structure our language and thought, even in regard to abstract topics. For instance, when talking about emotional support, we say \textit{to support a friend}.

\begin{figure}
    \centering
    \includegraphics[width=\columnwidth]{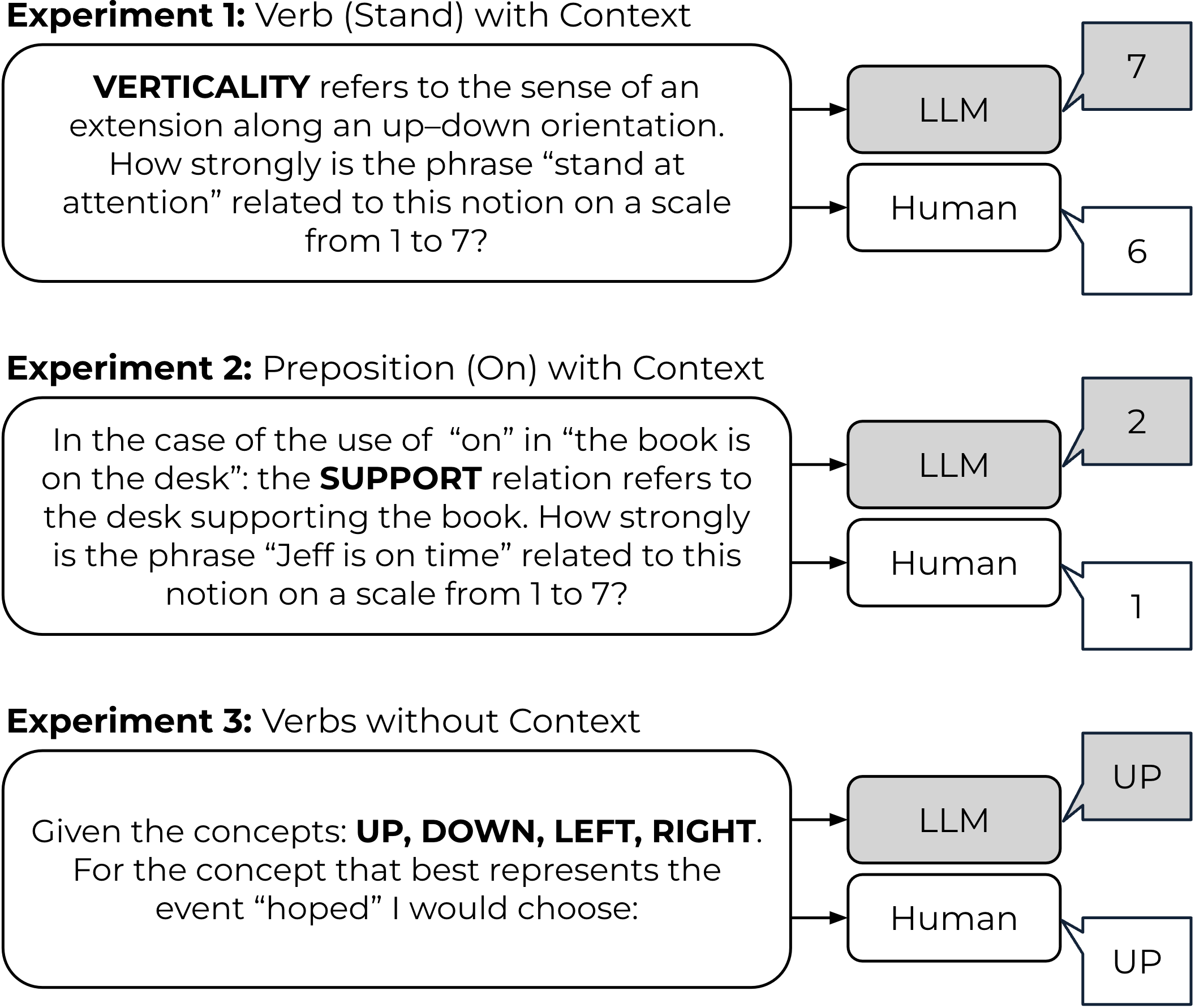}
    \vspace{-1em}
    \caption{Overview of the three experiments}
    \vspace{-1em}
    \label{fig:page1}
\end{figure}

Given that LLMs lack grounding, our research centers around a fundamental question: Can LLMs encode people's intuitions about the image schematic basis of words and phrases? In the subsequent sections, we present the rationale behind our inquiry and the methodologies for investigating the intricate relationship between language, embodiment, and the core aspects of human cognition. In summary, we make the following contributions:
\begin{itemize}
    \item We use LLMs and vision language models (VLMs) to reproduce \textbf{three psycholinguistics experiments} (Fig. \ref{fig:page1}) that connect language to humans' spatial intuitions
    \item We find that, in many instances, the answers of the largest models and of human participants \textbf{show moderate to strong correlations} even though the model's language use is not grounded in an embodied or enacted sense
    \item Crucial differences to human answers remain: the output of small and base models often shows low correlations, LLMs tend to give \textbf{polarized responses} (selecting either 1 or 7 on a scale), and VLMs show  lowest correlations with open-source VLMs showing none.
    
\end{itemize}

\section{Background}

\subsection{Image Schemas in Cognition \& Language}

Image schemas, rooted in cognitive linguistics and embodied cognition, have emerged as a foundational concept in our understanding of human language and cognition. These dynamic mental structures, originally proposed by \citet{lakoff2008metaphors}, serve as the cognitive building blocks for our conceptualization of the world. Image schemas are pre-linguistic and perceptual in nature, providing a fundamental means for humans to ground abstract concepts in concrete sensory and motor experiences \cite{johnson2013body,barsalou2008grounded}. This grounding is pivotal in comprehending and communicating complex ideas, as it bridges the gap between sensorimotor experiences and the vast array of abstract concepts that are integral to human thought and language. The centrality of image schemas in cognitive processes underscores their profound influence on language, from shaping our metaphors and linguistic expressions to facilitating our ability to reason, plan, and understand the world \cite{di2018linguistic,hampe2005image}.
An example of an image schema is \verticality. We can physically experience the image schema \verticality by standing upright or seeing one object positioned above another. In turn, these learned schemas can help us comprehend and communicate abstract concepts such as complex emotions (\textit{I feel down}) or power dynamics (\textit{she ranks high}).


Evidence for image schemas is provided by various psycholinguistic studies \cite{mandler1992build,gibbs1994taking,boroditsky2000metaphoric,richardson2001language,gibbs2005psychological}. For example, \citet{richardson2001language} has participants pick one of four arrows ($\uparrow,\downarrow,\leftarrow,\rightarrow$) that best represent a concrete or abstract action on horizontal or vertical dimension (e.g. concrete, horizontal: pushed / abstract, vertical: obeyed). The results indicate a common correlation that points towards underlying schemas in language. But how are these schematic intuitions represented by computational models of language?

\subsection{Image Schemas in Language Models}
LLMs are trained on vast amounts of text and code in order to construct a model of language that can be used for a variety of NLP tasks without having been specifically trained on these tasks \cite{radford2019language}. The ability of these emergent properties scales with the size of these models \cite{wei2022emergent}. Moreover, VLMs, e.g. GPT-4 \cite{openai2023gpt4}, are trained on images, text and code. Yet, neither LLMs nor VLMs are embodied systems in the sense that they never connect ``perception to action directly'' \cite{brooks1991intelligence}. This leads to the symbol grounding problem, as posed by \citet{harnad1990symbol}, questioning whether symbols can derive meaning just from other symbols alone (as would be the case for text-based LLMs) or whether they would need to be connected in a bottom-up fashion to sensory representations. To this day, such questions are being critically discussed in the NLP community \cite{bender-koller-2020-climbing}.
Research with embodied computational systems (e.g. robots) often works with the implicit premise (or challenge) that a system's physical form contributes to its technical capabilities, its affordances \cite{brohan2023can}. 
One way to look at the effect of embodiment in LLMs is provided by \citet{wicke2023lms}, who shows that the degree of perceived embodiment of an action word can have a positive effect on an LLM's capability to interpret figurative language. 


\section{Related Work}
\subsection{Human Behavioral Experiments with Large Language Models}
Using LLMs as human stand-in participants for psychology experiments has recently gained attention \cite{futrell2019neural, linzen2021syntactic, dillion2023can, harding2023ai, aher2023using}. Such use can be motivated by wanting to generate initial hypotheses for an experiment, pilot a new design, and gain insight into human cognition based on the assumption that LLMs trained on a large amount of human-generated text will produce similar output to that of human participants \cite{dillion2023can}. For instance, \citet{dillion2023can}, who propose such a use, report a high correlation of 0.95 between human answers and GPT-3.5 answers on moral judgment tasks. At the same time, they acknowledge that current LLMs are bad at capturing variation and diversity present in human responses and are biased towards responses of people from certain countries, economic backgrounds, genders, etc. 
\citet{harding2023ai} critique the use of LLMs to replace human participants and question the informativeness of the LLM's output.

Another motivation to simulate psychological experiments with LLMs is to gain insights not into human cognition but into the capabilities of language models themselves. Reproducing various experiments with LLMs, one can compare the LLM output with how humans behaved in the real experiment, thereby establishing the ``human-likeness'' of the model's text generations. The usefulness of such experiments has also been suggested with respect to psycholinguistics, where experiments can show what properties of language can be successfully processed, reproduced, or generated by LLMs \cite{houghton2023beyond}. 

A survey by \cite{linzen2021syntactic} presents studies of neural networks' syntactic abilities and their broader implications for linguistic theory.
\citet{dentella2023systematic}, for example, show that LLMs fail at distinguishing grammatical and ungrammatical sentences in a similar way to people. \citet{futrell2019neural} test four neural network language models on artificial sentences with syntactically complex structures (subordinate clauses and the Garden Path effect) to analyze their syntactic representations. Their findings indicate that LSTMs trained on large datasets represent syntactic states comparably to an RNNG trained on a small dataset, while an LSTM trained on a small dataset performs poorly or only weakly.
Other studies find that on many psychology tasks, the LLM output is comparable to human answers, even showing similar cognitive biases \cite{hagendorff2022machine, dasgupta2022language}. \citet{hagendorff2022machine} show that these cognitive biases tend to vanish when experimenting with the most recent models, such as ChatGPT and GPT-4.
\citet{aher2023using} extend the idea of repeating prominent experiments with LLMs. Specifically, they not only look at a single output of an LLM given some experiment prompt but try to simulate different demographics by prompting the model multiple times with different personas attached to each prompt. 
Generally, one needs to be aware that the same or similar experiments might have been in the training data.

\subsection{NLP for Image Schemas}
LLMs have only sparingly been used for image schema-related tasks. Initial research on the computational processing of image schemas was restricted to spectral cluster analysis \cite{gromann2017kinesthetic}, whereas more recent work \cite{wachowiak-etal-2022-drum, wachowiak-gromann-2022-systematic} uses the language model XLM-RoBERTa \cite{conneau-etal-2020-unsupervised}.  \citet{wachowiak-gromann-2022-systematic} show that language models can be fine-tuned to classify sentences based on eight image schema classes, with an accuracy between 57\% and 80\% depending on the language of the data.
Although their model requires first seeing more than 1,000 correctly annotated samples, these results indicate that, in principle, it is possible for a neural model trained only on text to pick up the pattern indicating a specific image schema in natural language. However, in contrast to the experiments in our work, the language samples they use are manually collated or created by experts. The psycho-linguistic experiments that build the foundation for the present work take a different approach, letting multiple people annotate phrases based on intuitions and felt relatedness. Accordingly, the expert-annotated image schemas are annotated with a single discrete label per sample, while the psycho-linguistic experiments inspiring this study lead to fine-grained annotations using ordinal scales for five different image schemas per phrase. 
\citet{kamath2023s} test various VLMs for whether they can correctly classify simple spatial configurations, such as A being left, right, under, or over B. Thus, they test for image schematic relations in their original physical form rather than the abstract extensions. They find models to perform poorly on the task, with the best model only achieving an accuracy of 60\%, while humans achieve 99\%, one of the reasons being that prepositions occur infrequently and in an ambiguous manner in the training data. Similar work by \citet{jassim2023grasp} shows shortcomings of VLMs' understanding of spatial configurations and intuitive physics when given different visual inputs of simulated spatial scenes. \\

From the corpus of related works delving into psycholinguistic studies probing spatial schema intuitions, we selected three experiments for comparison, each exhibiting variations in word class and context. Experiment 1 \cite{gibbs1994taking} showcases a verb (stand) within context, Experiment 2 \cite{beitel2001embodied} features a preposition (on) within context, and Experiment 3 \cite{richardson2001language} employs different verbs without context (see Fig. \ref{fig:page1}). Our rationale for excluding other studies stemmed from their failure to offer sufficient variation or to provide access to their original questionnaire. Moreover, the chosen experiments span a spectrum of distinct spatial schemas.

\section{Method}

This section provides an overview of the general model and prompt selection, while following subsections detail the setup for individual experiments.

\paragraph*{Model Selection.} 
We recreate the experiments using closed and open-source language models. The open-source LLaMA-2 instruction-tuned models \cite{touvron2023llama} are chosen in two sizes\footnote{Results of further experiments with non-instruct LLaMA models and their 7b version are included in the repository.}: LLaMA-2-13b and LLaMA-2-70b. The closed source models include: GPT-3 base (\texttt{davinci-002}), GPT-3.5 instruction-tuned (\texttt{text-davinci-003}) and GPT-4 (\texttt{gpt-4-0613}). Instruction-tuned versions have been chosen over base models because instruction-tuned models tend to perform better \cite{touvron2023llama}. This difference can be observed across all of our experiments in varying degrees. For the third experiment, we include three VLMs: the two open-source models IDEFICS-80b \cite{laurencon2023obelics} in base and instruction-tuned variants, as well as the GPT-4 Vision model (\texttt{gpt-4-vision-preview}).

\begin{table}
    \centering
    \scalebox{0.7}{
    \begin{tabular}{@{}ccccc@{}}
    \toprule
        \textbf{Model} & \textbf{Model}  & \textbf{Model} & \textbf{API} & \textbf{Open} \\
        \textbf{Type} & \textbf{Name}  & \textbf{Size} & \textbf{Endpoint} & \textbf{Source} \\
       \midrule
        LLM & GPT-3\textsubscript{base} & 175b & davinci-003  & -   \\
        LLM & GPT-3\textsubscript{inst} & N/A & text-davinci-003 & - \\
        LLM & GPT-4 & N/A & gpt-4-0613 & - \\
        
        LLM & LLaMA-2-13b & 13b & Llama-2-13b-chat-hf &  $\checkmark$ \\
        LLM & LLaMA-2-70b & 70b & Llama-2-70b-chat-hf &  $\checkmark$ \\
        &&&&\\
        VLM & GPT-4\textsubscript{vision} & N/A & gpt-4-vision-preview &  - \\
        VLM & IDEFICS-80b\textsubscript{base} & 80b & idefics-80b &  $\checkmark$ \\
        VLM & IDEFICS-80b\textsubscript{inst} & 80b & idefics-80b-instruct &  $\checkmark$ \\

    \bottomrule
    \end{tabular}
    }
    \caption{Models selected for experiments. LLM: large language model, VLM: vision--language model.}
    \label{tab:models}
\end{table}



\paragraph*{Prompt Selection.} \label{sec:method-prompt}Regarding their format, we keep the prompts as close to each original experiment as possible. Given the information provided in each paper, we reuse the wording of the image schema definitions and the items being evaluated. Besides these, we write our own instructive sentence that prompts the LLM to rate each item since the original instructions given to the human participants are unfortunately not provided in any of the papers. As suggested by \citet{aher2023using}, we optimize the instructive sentences by choosing a sentence that maximizes the fraction of valid model answers for each task. In experiments 1 and 2, valid answers consist of the numbers 1 to 7, while in experiment 3, the valid answers consist of the four possible directions. 
Given a set of valid answers $V$ and a prompt $k$, the validity score is computed as:
\begin{align}
\sum ^{}_{a \in V} p\left( a|k\right)
\end{align}
This way of finding a prompt allows us to get valid answers by only looking at their form but not at their content. Thus, we adopt this method that prevents overfitting caused by prompt-engineering. 

Recreating a prompt that closely mirrors the approach of the original paper would involve consolidating all the stimuli into a single list and instructing the models to rate each stimulus in combination with each image schema. We tried conditions in which the model had to rate all stimuli for all image schemas and all stimuli for a single schema. Additionally, we tried averaging multiple of those runs based on different stimuli orders. However, all preliminary experiments revealed that this comprehensive prompt yielded impractical or even unmeaningful responses. Frequently, models redundantly reproduced identical outputs for each item in the list. Consequently, we opted to refine our approach by providing the models with a single stimulus per input prompt.

\paragraph*{Evaluation.}
To evaluate how well the models can predict the human participant's judgment, we compute the Spearman correlation coefficient \cite{ca468a70-0be4-389a-b0b9-5dd1ff52b33f} between the human and the LLM ratings for each image schema per experiment. For interpretation, we use the labels weak, moderate ($>$0.4) and strong ($>$0.7), common in psychology literature \cite{akoglu2018user}.
In each of the three independent experiments, we addressed the challenge of multiple testing by employing the Benjamini-Hochberg correction for False Discovery Rate (FDR) \cite{benjamini1995controlling}. This correction was applied to account for the evaluation of multiple correlations within each experiment, ensuring a controlled rate of false positives and bolstering the statistical reliability of our findings. We use the \textit{statsmodels} python package \cite{seabold2010statsmodels} to evaluate the corrected p-values. All reported p-values are corrected and marked with * for $p<0.05$ as statistically significant.

\subsection{Experiment 1 - \citet{gibbs1994taking}}
\paragraph{The Original Experiment.} 

Firstly, we reproduce experiments by \citet{gibbs1994taking}, testing people's intuitions about the image schemas that underlie various uses of the verb \textit{to stand}. Given 32 phrases and the definitions of five relevant image schemas, they asked 27 participants to rate the relatedness between each image schema and each phrase on a Likert scale from 1 (``not at all related'') to 7 (``very strongly related''). 
The image schemas used in this experiment are: \balance, \verticality, \centerperiphery, \resistance, and \linkage. Participants have to rate the relatedness of all 32 phrases to a single image schema before moving to the next (we refer to this data as Gibbs\textsubscript{stand}). Before giving the rating for a particular image schema, the schema is introduced with a short definition. For instance, \verticality is introduced as referring ``to the sense of an extension along an up--down orientation''. 
The order in which the image schemas have to be rated is counterbalanced using five different orders overall. 
Additionally, Gibbs et al. repeat the experiment using the same 32 phrases but with \textit{stand} being replaced by a word with a synonymous meaning (we refer to this data as Gibbs\textsubscript{syn}). For example, ``to stand the test of time'' is substituted by ``to pass the test of time''.

\paragraph{The LLM Experiment.}



To extract ratings from an LLM, we retrieve the most likely answer generated, i.e. the answer received with a \textit{temperature} of 0 in the OpenAI API or a \textit{top\_k} of 1 with HuggingFace.  
Alternatively, one could consider the probability for each valid answer, the numbers between 1 and 7, and compute the sum of each number weighted by their likelihood, normalized by the sum of all seven numbers' likelihoods. While this takes the LLM's uncertainty into account, it also requires seven times the amount of compute compared to simply taking the top answer. As results only varied minimally, we chose the cost-effective methodology of selecting only the top answer.

\paragraph*{Prompting.}
For experimenting, we started with a basic input text as close as possible to the original experiment, for example:
\begin{quote}
\small{
Consider the notion of \verticality. Verticality refers to the sense of an extension along an up--down orientation. How strongly is the phrase ``stand at attention'' related to this notion on a scale from 1 (not at all related) to 7 (very strongly related)?
}
\end{quote}
The different image schema definitions used in these prompts can be found in Appendix \ref{tab:exp1_is_defs}.
Given such a text as the start of the input, we try to find a way to end the prompt so that the model's output probabilities for the next token converge towards 100\% when summed for all valid 7 numbers. This final bit of the prompt depends on the model. Our prompt ending choices are described in Section \ref{ref:res_exp1}, and all used phrases are listed in the Appendix \ref{tab:exp12_items}.


\subsection{Experiment 2 - \citet{beitel2001embodied}}
\paragraph{The Original Experiment.} \citet{beitel2001embodied} repeat the experimental paradigm established by \citet{gibbs1994taking}, however, with a new set of phrases, all containing the preposition \textit{on}. Given the focus on the word \textit{on} instead of \textit{to stand}, they also select a different set of image schemas that are more relevant in this case, namely: \support, \pressure, \constraint, \covering, \visibility. Instead of having access to a general definition of each image schema, the participants can now check an example sentence for which five introduction statements explain how each image schema relates to it. For example, the experiment introduces \support by saying, ``In the case of the use of ``on'' in the ``the book is on the desk'': the \support relation refers to the desk supporting the book''. A list of introductions for all five image schemas is available in the Appendix Table \ref{tab:exp2_is_defs}.

\paragraph{The LLM Experiment.} Given its identical structure, the experiment is conducted in the same manner as with the data from \citet{gibbs1994taking}.

\subsection{Experiment 3 - \citet{richardson2001language}}

\begin{figure}
    \centering
    \includegraphics[width=.475\textwidth]{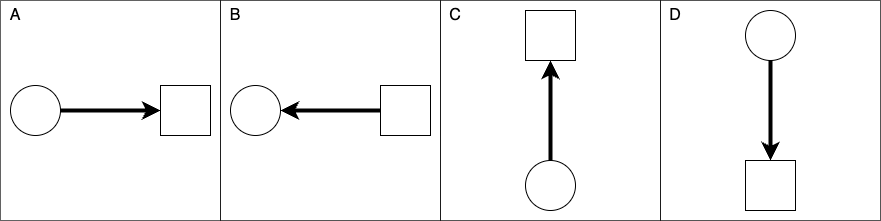}
    \caption{Target images form the original study by \citet{richardson2001language}. Each participant was asked to match 30 verbs to one of the images (A-D).}
    \label{fig:ogexp1}
\end{figure}

\paragraph{The Original Experiment.} 
\citet{richardson2001language} provide experimental evidence for image schemas via two different experiments. The first experiment presents human participants with lists of 30 verbs ranked by concreteness (based on the MRC
psycholinguistic database \cite{coltheart1981mrc}). The verbs are represented using an agent-patient relation through a circle-square depiction, e.g. $\bigcircle$ \textit{offended} $\square$. Distinguished by abstractness versus concreteness of the verbs, those verbs were further divided into three groups based on their primary direction: horizontal, vertical and neutral. This resulted in a 2x3 factor design of concreteness by directionality. The task for each participant was to choose one of four images that best represents the action. The images use the same circle-square relation, but with an arrow representing the directionality on the horizontal and vertical axis ($\leftarrow,\rightarrow,\uparrow,\downarrow$). The choice of images is depicted in Figure \ref{fig:ogexp1}. As an example, for the horizontal, concrete item $\bigcircle$ \textit{offended} $\square$ the participant needs to select one of the A-D images (Fig. \ref{fig:ogexp1}). It is important to note that \citet{richardson2001language} analyse the results with respect to the primary direction (horizontal/vertical). The second experiment by \citet{richardson2001language} requires participants to draw a schematic representation of items from the first experiment. Since we do not include visual generation, we do reproduce the second experiment in the presented work.

\paragraph{The LLM Experiment.}  \label{sec:method_gibbs}

The original experiment involved the use of four visual depictions (see Fig.\ref{fig:ogexp1}) alongside a list of verbs. In this setup, participants were exposed to all the images and words simultaneously. The word presentation order was randomised, and the images were labeled A-D. To replicate the experiment, which necessitates a visual input, we also employ VLMs. In order to enable comparisons with our selected (text-only) LLMs, we design a tripartite experimentation.

In the first phase, we opted to translate the visual depictions into \textbf{textual} representations of the underlying schemas, specifically the words \texttt{up}, \texttt{down}, \texttt{left}, and \texttt{right}. For the second phase, we conducted the experiment again, but this time with \textbf{pseudo-visual} renderings using Unicode arrows (↑, ↓, ←, →). In the final phase, we turned to a limited selection of VLMs to rerun the experiment with textual and \textbf{visual} input, showing the actual images from Fig. \ref{fig:ogexp1}. This tripartite approach allows us to investigate the impact of visual input on language models, compare it to text-only models, and explore the utility of different forms of visual and textual representations across three conditions.

Importantly, we compare the model responses with the results from the \citet{richardson2001language} experiment. Initially, our experiments included the analysis of left, right, up and down decisions, which require an additional encoding of the agent-patient relation through a circle-square depiction. This encoding diffused any correlations and we identified that a restrictions to primary directions (horizontal/vertical) by grouping up/down and left/right options in the analysis yields a better reflection of the the models correlations, since it does not require an agent-patient encoding.

\paragraph{Prompting.} 

Detailed lists of prompts for each condition can be found in Appendix Table \ref{tab:exp3_prompts}. 
We observed a significant influence on the model's response based on the order in which the four options were presented, whether as concept words, Unicode arrows or images. This is in-line with findings by \citet{pezeshkpour2023large}. Regardless of different option orders (e.g., ↑, ↓, ←, →), the model most often favoured the first option (e.g., ↑) as a default answer for most items. To mitigate this effect and to acquire a distribution for each verb, we conducted all 24 possible permutations of the choices per word (4! permutations of 30 verbs).

VLMs require a label attached to each image, which allows them to formulate a choice, e.g. ``Image A''. We observed that labels with an implicit (alphabetical or numerical) order, e.g. A, B, C or 1, 2, 3, introduce a selection bias. Hence, we use arbitrary labels (VMBR, WJZX, XQHL, YGPK).

Lastly, the GPT-4 models are a commercial product that include safety guardrails in their system prompt in order to avoid hate speech, abuse or disinformation. Often, GPT-4 models would tend to refuse to answer subjective questions. Consequently, we modified instructions for GPT-4 in order to force subjective model answers (Fig. \ref{tab:exp3_prompts}).

\section{Results}

\renewcommand{\arraystretch}{1.3}
\begin{table*}[t]
\centering{ 
\begin{tabular}{@{}llllllll@{}}
\toprule
    \textbf{Exp.} & \textbf{Image Schema}& \multicolumn{2}{c}{\textbf{LLaMA-2}} & \multicolumn{3}{c}{\textbf{GPT}} & \textbf{Avg.}\\
\cmidrule(l){3-4}
\cmidrule(l){5-7}
    & &  \textbf{13b-chat} & \textbf{70b-chat}& \textbf{GPT-3\textsubscript{base}} & \textbf{GPT-3\textsubscript{inst}} & \textbf{GPT-4}\\
    
    \hline
        \multirow{5}{*}{Gibbs\textsubscript{stand}}
        &\verticality&0.26&0.41*&N/A&0.53*&\textbf{0.69*} & 0.47\\
        &\balance&0.27&0.38&-0.05&0.37&\textbf{0.49*} & 0.29\\
        &\centerperiphery&0.20&0.36&N/A&\textbf{0.82*}&0.56* & 0.49\\
        &\linkage&0.24&-0.06&N/A&0.46*&\textbf{0.61*} & 0.31\\
        &\resistance&0.41*&0.48*&N/A&0.71*&\textbf{0.82*} & 0.60\\
        &Avg.&0.28&0.31&N/A&0.58&\textbf{0.63 }& \\
        
    \hline
        \multirow{5}{*}{Gibbs\textsubscript{syn}}
        &\verticality&0.22&0.57*&N/A&0.49*&\textbf{0.70*} & 0.49\\
        &\balance&0.36&0.50*&0.17&0.50*&\textbf{0.54*} & 0.41\\
        &\centerperiphery&0.22&0.32&0.16&\textbf{0.67*}&0.67* & 0.41\\
        &\linkage&0.61*&0.32&N/A&\textbf{0.61*}&0.24 & 0.45\\
        &\resistance&0.30&0.54*&N/A&0.61*&\textbf{0.77*} & 0.56\\
        &Avg. &0.34&0.45&N/A&\textbf{0.61}&0.58 & \\

    \hline
        \multirow{5}{*}{Beitel\textsubscript{on}}
        &\support&0.19&0.32&N/A&0.48*&\textbf{0.62*} & 0.40\\
        &\pressure&0.37*&0.72*&N/A&\textbf{0.79*}&0.37* & 0.56\\
        &\constraint&0.49*&0.37*&N/A& \textbf{0.60*}&0.47* & 0.48\\
        &\covering&-0.15&0.41*&N/A&0.46*&\textbf{0.68*} & 0.35\\
        &\visibility&0.24&0.38*&N/A& \textbf{0.69*} &0.62* & 0.48\\
        &Avg.&0.23&0.44&N/A&\textbf{0.60}& 0.55& \\

\bottomrule
\end{tabular}}
\caption{Spearman correlation: model answers and human answers. * for $p<0.05$, \textbf{bold} = highest correlation} \label{tab:spearman}
\end{table*}
\renewcommand{\arraystretch}{1}

\subsection{Experiment 1 — \citet{gibbs1994taking}} \label{ref:res_exp1}
\paragraph{Prompt Selection.} 
Depending on the chosen model, different prompt endings worked well in making the respective LLM generate valid answers, that is, a number between 1 and 7. After considering the output probabilities over a subset of 15 input samples, we ended up using either prompt ending ``Only answer with the score:'' or ``I choose the number''. In each case, the chosen prompt ending guarantees that more than 99\% of the probability mass is allocated to the valid answers. For GPT-4, the log-probabilities are not accessible through the API, which is why we reused the prompt ending chosen for davinci-003. Based on the 15 tested samples, we also assessed whether the models are very sensitive to these scores in relation to the number they output. Fortunately, we found that the scores are relatively stable and, on average, differ only around half a point on the 7-point scale.  


\paragraph{Answer Correlations.}

First, we compare the two evaluation methods presented in Section \ref{sec:method_gibbs}, i.e., simply extracting the most likely answer or computing the average of all valid answers based on their likelihood. Using davinci-003, we find that, on average, the thereby extracted answers are only 0.39 points apart. Thus, we decided to use the simpler method of using the most likely answer when presenting the results from hereinafter since it has the advantage of incurring fewer computational costs. 

Table \ref{tab:spearman} shows the Spearman correlations between human and model answers for the five different image schemas. For 70\% of the image schemas, GPT-4 generates the answers most similar to those of human participants, with GPT-3\textsubscript{inst} generating the most similar answers for the remaining 30\%. On average, GPT-4 shows a correlation of 0.61, which can be interpreted as moderate. 
In comparison, the LLaMA-2 models produce answers that are more dissimilar to those of humans, as indicated by some moderate correlations and many correlations that are not statistically significant. The 13b variant achieves an average correlation of 0.31, and the 70b variant of 0.38. These lower correlations can be partly attributed to the fact that the LLaMA-2 models tend to answer with either 4 or 7, largely ignoring other options on the scale. 
The only model that is trained without reinforcement learning from human feedback, GPT-3\textsubscript{base}, fails to generate insightful answers, nearly always generating a 4. 

Despite GPT-4 showing the highest correlations, it still generates some answers strongly deviating from human spatial intuitions. For example, GPT-4 fails to relate a sense of VERTICALITY to the phrase ``the barometer stands at 30 inches'' (scoring it 1 compared to the participant average of 4.71) or BALANCE to the phrase ``the clock stands on the mantle'' (1 compared to 4.46). Generally, it can be observed that many of the outliers are caused when the model gives too low ratings, i.e., 1 or 2.



\subsection{Experiment 2 --- \citet{beitel2001embodied}}
Results are similar to the previous experiment, with the answers of GPT-4 and GPT-3\textsubscript{instruct} showing the highest similarity to those of human participants, ranging between correlations of 0.4 and 0.8. 

\subsection{Experiment 3 --- \citet{richardson2001language}}

\renewcommand{\arraystretch}{1.3}
\begin{table*}[!t]
\centering{ 
\scalebox{0.8}{
\begin{tabular}{@{}lllllllllc@{}}
    \textsc{TEXT} & \textbf{Llama-2-13b\textsubscript{chat}} & \textbf{Llama-2-70b\textsubscript{chat}} & \textbf{GPT-3\textsubscript{base}} & \textbf{GPT-3\textsubscript{inst}} & \textbf{GPT-4} & \textbf{Avg.} \\

    \hline    
        \textsc{Horiz.} / \textsc{Vert.} &
        0.53* / 0.54* & 0.59* / 0.59* & 0.68* / 0.68* & \textbf{0.72}* / \textbf{0.72}* & 0.58* / 0.58* & 0.57 / 0.59
        \\

\cmidrule(l){2-7}

    \textsc{PSEUDO-VISUAL} & \textbf{Llama-2-13b\textsubscript{chat}} & \textbf{Llama-2-70b\textsubscript{chat}} & \textbf{GPT-3\textsubscript{base}} & \textbf{GPT-3\textsubscript{inst}} & \textbf{GPT-4} & \textbf{Avg.} \\

    \hline    
        \textsc{Horiz.} / \textsc{Vert.} &
        0.68* / 0.79* & 0.51* / 0.51* & 0.17 / 0.16 & 0.68* / 0.68* & \textbf{0.82}* / \textbf{0.83}* & 0.61 / 0.66
        \\
\hline

        Avg. &
        0.61 / 0.67 & 0.55 / 0.55 & 0.43 / 0.42 & \textbf{0.70} / 0.70 & \textbf{0.70} / \textbf{0.71} &  \\

\\

    \textsc{VISUAL} & \textbf{IDEFICS-80b\textsubscript{base}} & \textbf{IDEFICS-80b\textsubscript{inst}} & \textbf{GPT-4\textsubscript{vision}} &  & & \textbf{Avg.}  \\

    \hline    
        \textsc{Horiz.} / \textsc{Vert.} &
        0.00 / 0.00 & -0.01 / -0.01 & \textbf{0.57}* / \textbf{0.56}* & & & 0.19 / 0.18
        \\

\end{tabular}}}
\caption{Spearman correlation: model answers and human answers. * for $p<0.05$, \textbf{bold} = highest correlation} 
\label{tab:exp3_res}
\end{table*}
\renewcommand{\arraystretch}{1}

\paragraph{Prompt Selection.} 
In contrast to the previous experiments, the model output is not an ordinal measure (1-7), but a nominal classification. Analogously, we use a subset of input samples along with various different prompt endings to identify that the chosen prompt ending guarantees that a high probability mass is allocated to the labels.

Notable results can be reported for text and vision prompting. For text-based models, the addition of quotation marks around the label (e.g. \textquotesingle up\textquotesingle) increases the probability of the model to choose a valid label if prompted with a quote at the end. In general, using a ``Question:'' and ``Answer:'' structure improves label likelihood. For VLMs we cannot obtain the log-probabilities, therefore we follow the examples provided by the model developers\footnote{\url{https://huggingface.co/docs/transformers/main/en/tasks/idefics}}.

\paragraph{Results of Primary Directionality.} For all models and all conditions, we summarise the Spearman correlations in Table \ref{tab:exp3_res} (detailed results for all choices are listed in the Appendix Tab. \ref{tab:exp3_all}). We can observe strong correlations (> 0.7) in the textual and pseudo-visual conditions, but not in the visual one. In the textual condition, all models show a significant correlation with the human choices with respect to their choice of a primary directionality (horizontal/vertical). Here, GPT-3\textsubscript{inst} shows a strong correlation of 0.72. Correlations for pseudo-visual conditions are higher except for one outlier: GPT-3\textsubscript{base}, which has mostly selected the ↑ for all items irrespective of the order of choices (i.e. order of arrows in the prompt). On the contrary, GPT-4 has the highest correlation across all models and conditions for the pseudo-visual task with 0.82 (horizontal) and 0.83 (vertical). In the visual condition, both open-source versions of IDEFICS do not show any correlation with the human responses. GPT-4\textsubscript{vision} achieves a significant, but moderate correlation with 0.57 (horizontal) and 0.56 (vertical). On average, correlations between model and human answers are higher in the pseudo-visual condition despite the outlier of the GPT-3\textsubscript{base} model.

\section{Discussion}
\begin{figure*}
    \centering
    \includegraphics[width=\textwidth]{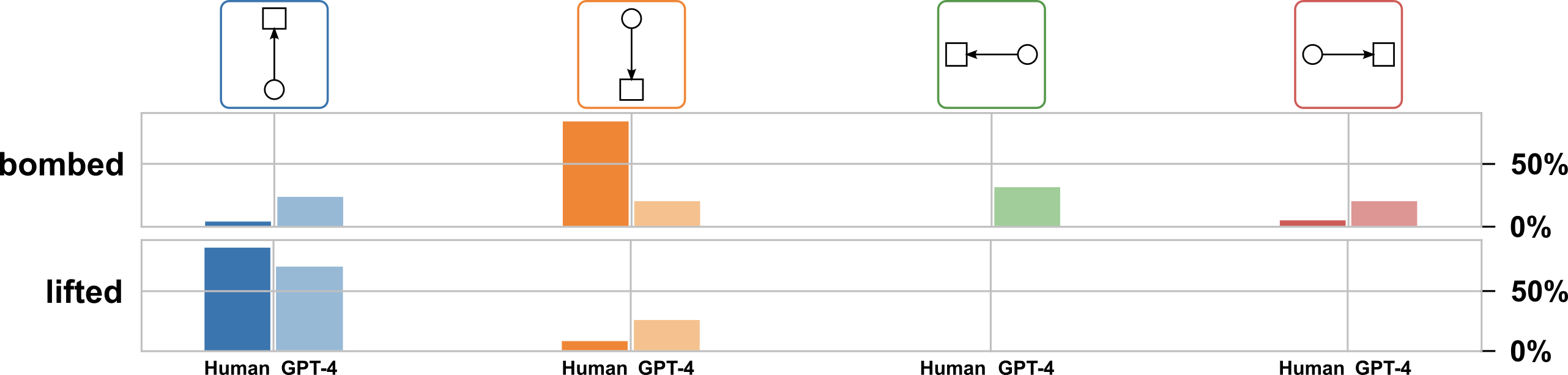}
    \caption{Distribution of image schema choice for items ``bombed'' and ``lifted'' by humans (bold) and GPT-4 (light).}
    \label{fig:results1}
\end{figure*}



We explored if LLMs capture human intuitions about image schematic basis in language through three psycholinguistic experiments comparing LLMs and VLMs. Model responses often correlate with people's, especially in larger models, although discrepancies exist for certain image schemas. Nevertheless, the models reflect spatial primitive intuitions, potentially stemming from their ability to model words, their contextual use, and their relation to schema definitions. Another possibility is that the original experiment papers, serving as training data, might contribute to the models' reproduction of observed patterns, although parsing the original papers' results effectively is unlikely.

At the same time, one might wonder why, in some cases, the model answers are so far apart from human answers.
Besides the explanation of the lack of embodied experience, one aspect to consider is that the LLMs only had access to one item at a time when rating stimuli. Thus, models are unable to rate items relative to each other --- a strategy a human participant is likely to adopt. For instance, given the two stimuli ``stand in awe'' and ``the clock stands on the mantle'' separately, GPT-4 gives scores of 2 and 1 for the image schema \balance. People, on the other hand, tend to relate \balance much more strongly to a clock standing on a mantle. This intuition is captured when prompting the model with both stimuli at once, thus allowing the model to provide relative scores; in this case, the model provides a score of 2 for the first phrase but a score of 4 for the clock on the mantel, thus being much closer to human scores. However, prompting the model to provide all answers at once is currently not feasible, as described in Section \ref{sec:method-prompt}.
This missing comparison between items is also reflected in the fact that the models tend to use the extremes of the scale, i.e., rating items with 1 or 7. 
Another possible answer for the differences is related to the partially small participant pool used in the original experiments. The three experiments recruited 27, 24, and 173 participants. Therefore, the originally recorded human answers might not be very robust, and future experiments should rerun similar setups not only with novel stimuli but also with larger participant pools to enable a more robust data set for comparison.

The overall weakest correlations are observed for VLMs, particularly with open-source model IDEFICS, displaying meaningless responses. For instance, IDEFICS-80b\textsubscript{inst} consistently selects the left-arrow image, disregarding the item, image order, or randomized labels. This failure is attributed to VLMs being trained on natural images, unsuitable for interpreting the highly abstract line drawings. Even GPT-4\textsubscript{vision}, despite showing moderate correlations (0.56-0.57), occasionally provides random answers, deviating from human consensus (see Fig. \ref{fig:results1}). This discrepancy raises questions about the models' alignment with human intuitions.



\section{Conclusion}
This exploratory study is the first of its kind to reproduce psycholinguistic experiments in order to explore spatial schema intuitions. Moreover, it provides evidence that LLMs are able to reflect those intuitions in different tasks and setups. The results also point out that despite the duo-modality, VLMs do not encode the spatial understanding as effectively as their textual origins.
Future research aims to assess models using novel, collected stimuli, ensuring no prior exposure during training. Our current findings provide a foundation for formulating precise, testable hypotheses in subsequent experiments. Additionally, we would like to extend this line of research to include a multilingual analysis. In conclusion, our study not only sheds light on the disparities between LLMs and human cognition but also paves the way for new research perspectives in understanding and refining language models. By providing empirical evidence of these disparities, we advocate for a deeper exploration into the limitations of current models and the development of novel approaches to bridge the gap between artificial intelligence and human intelligence.

\section{Limitations}

While our investigation into the applications of proprietary models, such as GPT-3 and GPT-4, offers valuable insights, it is essential to acknowledge the inherent limitations associated with their use. The opaque nature of the underlying mechanisms in proprietary models poses a significant limitation. The understanding is constrained by the lack of detailed information on the architectural intricacies, leaving us to make assumptions based on analogies to open-source models. It is crucial to recognize that variations in architecture, data, and parameters between proprietary and open-source models impact the generalizability of our findings. \\

Furthermore, the original experiments have a limited demographic, which we hereby report: \cite{gibbs1994taking} with 27 undergraduate students, U.S. university, native English speakers. \cite{beitel2001embodied} with 24 undergraduate students, U.S. university, native English speakers. \cite{richardson2001language} with 173 undergraduate students, U.S. university, no further information is provided. 

Additionally, our reliance on psycholinguistic data introduces notable limitations. Firstly, the temporal aspect of the data is a concern, given its age range of 30 to 23 years. Language evolves over time, and the potential disparities between our data and contemporary linguistic trends may affect the applicability of our results. Therefore, we plan to replicate the studies to gauge temporal robustness. Moreover, we acknowledge that the reproduced studies solely feature the English language and multilingual analysis is subject of future work.

Secondly, the incorporation of original papers into the training data of proprietary models, particularly the LLM and VLM, poses challenges. This integration may introduce biases, potentially influencing the outcomes of our experiments. Yet, the original papers' results are presented in formats that are unlikely to have undergone parsing during the training procedures, especially within the context of the VLM, yet this remains hard to prove.

\section{Considerations and Impact of the Work}

\paragraph{Environmental Impact}
Overall, text generations with the OpenAI API cost 25.85\$. Text generations with LLaMA models were run on two clusters. In those experiments, we utilized 8 NVIDIA RTX A6000 (48GB) GPUs for a 4-hour runtime, each with a power consumption of 300 W, resulting in an estimated total power consumption of 9.6 kWh and a CO2 emission of approximately 3.984 kg. Additionally, we used 2 NVIDIA A100 (40GB) GPUs for a 39-hour runtime, also consuming 300~W each, contributing to a combined total power consumption of 33 kWh and a CO2 emission of ˜13.7 kg.

\section*{Acknowledgements}
 Lennart Wachowiak was supported by UK Research and Innovation (EP/S023356/1), in the UKRI Centre for Doctoral Training in Safe and Trusted Artificial Intelligence, as well as the King's Institute for Artificial Intelligence.

\bibliography{anthology, custom}
\bibliographystyle{acl_natbib}

\appendix
\label{appendix}

\setcounter{table}{0}
\renewcommand{\thetable}{A\arabic{table}}

\renewcommand{\arraystretch}{1.3}
\begin{table*}[ht]
\centering{
\captionsetup{justification=raggedright,singlelinecheck=off}
\caption*{\Large{\textbf{Appendix}}}

\begin{tabular}{@{}l|p{12cm}@{}}
\toprule
\textbf{Image Schema} & \textbf{Definition/Introduction} \\
\midrule 
\balance & Consider the notion of \balance. Balance refers to your sense of symmetry or stability relative to some point within your body. \\
\verticality & Consider the notion of \verticality. Verticality refers to the sense of an extension along an up--down orientation.
\\
\centerperiphery &  Consider the notion of \centerperiphery. Center--periphery refers to the experience of some objects or events as central while surrounding objects and events are peripheral or to the outside. \\
\resistance & Consider the notion of \resistance. Resistance refers to the experience of your body opposing some external force. \\
\linkage& Consider the notion of \linkage. Linkage refers to the perception of a connection between objects or events. \\
\bottomrule
\end{tabular}}
\caption{Image schema definitions provided in the prompts (Experiment 1)}\label{tab:exp1_is_defs}
\end{table*}
\renewcommand{\arraystretch}{1}

\renewcommand{\arraystretch}{1.3}
\begin{table*}[ht]
\centering{

\begin{tabular}{@{}l|p{13cm}@{}}
\toprule
\textbf{Image Schema} & \textbf{Definition/Introduction} \\
\midrule 
\support &  In the case of the use of ``on'' in ``the book is on the desk'': the \support relation refers to the desk supporting the book.\\
\pressure &  In the case of the use of ``on'' in ``the book is on the desk'': the \pressure relation refers to the book exerting some pressure on the desk.\\
\constraint &   In the case of the use of ``on'' in ``the book is on the desk'':  the \constraint relation refers to the desk constraining the possible motions of the book.\\
\covering &  In the case of the use of ``on'' in ``the book is on the desk'': the \covering relation refers to the book concealing the part of the desk that is under the book.\\
\visibility& In the case of the use of ``on'' in ``the book is on the desk'': the \visibility relation refers to the book being visible on the desk.\\
\bottomrule
\end{tabular}}
\caption{Image schema definitions provided in the prompts (Experiment 2)}\label{tab:exp2_is_defs}
\end{table*}
\renewcommand{\arraystretch}{1}

\renewcommand{\arraystretch}{1.0}
\begin{table*}[ht]
\centering{
\scalebox{0.7}{
\begin{tabular}{@{}l|l|l@{}}
\toprule
\textbf{Stimuli with stand (Experiment 1)} & \textbf{Stimuli with synonym (Experiment 1)} & \textbf{Stimuli with on (Experiment 2)} \\
\midrule 

stand at attention& be at attention& The family depends on the father\\
stand out in several sports& be distinguished in several sports& There is a physician on call\\
to stand firm& to hold firm& All books are on sale\\
don't stand for such treatment& don't allow such treatment& The band is on tour\\
to stand the test of time& to pass the test of time& The boat is on course\\
united we stand& united we are strong& The bus is on schedule\\
we stand on 30 years of experience& we are backed up by 30 years experience& Jeff is on time\\
let the issue stand& let the issue remain as is& Sam is on his way home\\
let the mixture stand& leave the mixture undisturbed& She puts the blame on my actions\\
get stood up for a date& have a date with someone who didn't show up& He pulled a gun on me\\
he stands six-foot nine& he measures six-foot nine& Pat has been on sick leave\\
the clock stands on the mantle& the clock is on the mantle& There is a parade on Sunday\\
one-night stand& one-night fling& The program will be broadcast on CBS\\
to stand to profit& to be in the position to make a profit& Joan works on the committee\\
to stand in someone else's shoes& to be in someone else's shoes& Linda is very knowledgeable on this subject\\
on the witness stand& on the witness platform& The factory workers are on strike for a second day\\
stand in awe& be in aw& These cold nights are very hard on the homeless\\
the police told them to stand back& the police told them to get back& They ordered a court martial to be held on him\\
stand by your man& support your man& Pam is on a diet\\
the engine can't stand the constant wear& the engine can't endure the constant wear& The dog is on the leash\\
stand on shaky ground& to be on shaky ground& She is on the pill\\
to stand accused& to be accused& The ship is on the anchor\\
the house stands in the clearing& the house is in the clearing& The lunch is on George\\
the barometer stands at 30 inches& the barometer is at 30 inches& He lives on a pension\\
as the matter now stands& as the matter now exists& The boss is on my neck\\
the part stands for the whole& the part represents the whole& There are fifty nurses on the hospital staff\\
it stands to reason& it conforms with reason& There is ten cent interest on the dollar\\
they did nothing but stand around& they did nothing but hang around& The vase is on the table\\
to stand against great odds& to face great odds& The fish is on the surface\\
a standing ovation& a roaring ovation& The fly is on the ceiling\\
the boss always stands over me& the boss always hovers over me& She has a beautiful ring on her finger\\
he stands committed& he remains committed& The house rests on the foundation\\
& & They hung a picture on the wall\\
& & I have a cut on my finger\\
& & I have a rash on my back\\
& & The actor is on stage\\
& & He resides on the continent\\

\bottomrule
\end{tabular}}
}
\caption{Items the LLMs had to rate in experiments 1 and 2}\label{tab:exp12_items}
\end{table*}
\renewcommand{\arraystretch}{1}

\begin{table*}
\label{tab:word_list}
\centering
\begin{tabular}{p{\textwidth}}
\toprule
\textbf{Action words (Experiment 3)} \\
\midrule
fled, pointed at, pulled, pushed, walked, hunted, impacted, perched, showed, smashed, bombed, flew, floated, lifted, sank, argued with, gave to, offended, rushed, warned, owned, regretted, rested, tempted, wanted, hoped, increased, obeyed, respected, succeeded \\
\bottomrule
\caption{List of action words from \cite{richardson2001language}. These words are used for Experiment 3.}
\end{tabular}
\end{table*}

\renewcommand{\arraystretch}{1.3}
\begin{table*}[ht]
\centering{
\begin{tabular}{@{}l|p{12.33cm}@{}}
\toprule
\textbf{Condition} & \textbf{Prompt} \\
\midrule 
\textsc{Textual} & Given the concepts: \texttt{[DIRECTION WORDS]}. For the concept that best represents the event \textquotesingle\texttt{[ACTION WORD]}\textquotesingle, I would choose \textquotesingle
\\
\cmidrule(l){2-2}
\textsc{Pseudo-Visual} & Given the event \textquotesingle\texttt{[ACTION WORD]}\textquotesingle, which of the following arrows best represents this event: \texttt{[UNICODE ARROWS]}. A research participant would choose the arrow: 
\\
\cmidrule(l){2-2}
\textsc{Visual} & 
\texttt{[}Instruction: You are shown four images. Select one of the four images to answer the question.\texttt{]}

Image \texttt{[IMAGE LABEL]}:\texttt{[IMAGE]}

Image \texttt{[IMAGE LABEL]}:\texttt{[IMAGE]}

Image \texttt{[IMAGE LABEL]}:\texttt{[IMAGE]}

Image \texttt{[IMAGE LABEL]}:\texttt{[IMAGE]}

Question: Which is the best image to describe the event \textquotesingle\texttt{[ACTION WORD]}\textquotesingle? Answer: For \textquotesingle\texttt{[ACTION WORD]}\textquotesingle \hspace{0.2 em}it is Image  \\ 
\cmidrule(l){2-2}
\textsc{GPT-4\textsubscript{chat}} & \texttt{[TEXTUAL} or \texttt{PSEUDO-VISUAL PROMPT]} You are a participant in a research experiment. Even if the answer is subjective, provide it. Do not say it is subjective. 
\\
\bottomrule
\end{tabular}}
\caption{Prompts of the three different modality-conditions used in Experiment 3. \texttt{DIRECTION WORDS}: \textquotesingle up\textquotesingle, \textquotesingle down\textquotesingle, \textquotesingle left\textquotesingle, \textquotesingle right\textquotesingle. \texttt{UNICODE ARROWS}: ↑, ↓, ←, →. \texttt{IMAGE LABEL}: VMBR, WJZX, XQHL, YGPK. \texttt{IMAGE} see Fig. 1 (without letter labels). 
}\label{tab:exp3_prompts}
\end{table*}
\renewcommand{\arraystretch}{1}

\renewcommand{\arraystretch}{1.2}
\begin{table*}[t]
\centering{ 
\begin{tabular}{@{}lllllllc@{}}
\toprule
    \textbf{Condition} & \textbf{Image Schema}& \multicolumn{2}{c}{\textbf{LLaMA-2}} & \multicolumn{3}{c}{\textbf{GPT}} & \textbf{Avg.}\\
\cmidrule(l){3-4}
\cmidrule(l){5-7}
    & &  \textbf{13b-chat} & \textbf{70b-chat}& \textbf{GPT-3\textsubscript{base}} & \textbf{GPT-3\textsubscript{inst}} & \textbf{GPT-4} &\\

    \hline
        \multirow{4}{*}{\textsc{TEXTUAL}}
        &\textsc{up}&0.49&0.66&\textbf{0.67}&0.63&0.66 & 0.62\\
        &\textsc{down}&0.21&0.23&\textbf{0.34}&0.31&0.33& 0.28\\
        &\textsc{left}&-0.12&0.25&0.23&\textbf{0.37}&0.24& 0.24\\
        &\textsc{right}&\textbf{0.57}&0.56&0.53&0.56&0.41& 0.53\\

    \hline
        \multirow{4}{*}{\textsc{PSEUDO-VISUAL}}
        &$\uparrow$&0.44&0.61&-0.09&0.49&\textbf{0.70}& 0.43\\
        &$\downarrow$&0.44&0.14&0.31&0.42&\textbf{0.49}& 0.36\\
        &$\leftarrow$&0.18&0.29&0.03&\textbf{0.31}&0.18& 0.20\\
        &$\rightarrow$&0.31&0.43&N/A&0.56&\textbf{0.69}& 0.50\\

\bottomrule
\end{tabular}}
\caption{Spearman correlation between model answers and human answers.  \textbf{bold} = highest correlation}
\label{tab:exp3_all}
\end{table*}
\renewcommand{\arraystretch}{1}

\renewcommand{\arraystretch}{1.2}
\begin{table*}[t]
\centering{ 
\begin{tabular}{@{}lllllc@{}}
\toprule
    \textbf{} & \textbf{Image}& \multicolumn{2}{c}{\textbf{IDEFICS}} & \multicolumn{1}{c}{\textbf{GPT}} & \textbf{Avg.}\\
\cmidrule(l){3-4}
\cmidrule(l){5-5}
    & &  \textbf{80b} & \textbf{80b-inst}& \textbf{GPT-4} &\\

    \hline
        \multirow{4}{*}{\raisebox{-5 em}{\textsc{VISUAL}}}
        &{\includegraphics[width=15mm, height=15mm]{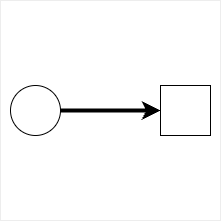}}&
        \raisebox{1.5 em}{0.12}&
        \raisebox{1.5 em}{0.10}&
        \raisebox{1.5 em}{\textbf{0.51}}&
        \raisebox{1.5 em}{0.24}\\
        
        &{\includegraphics[width=15mm, height=15mm]{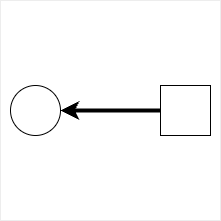}}&
        \raisebox{1.5 em}{-0.12}&
        \raisebox{1.5 em}{0.01}&
        \raisebox{1.5 em}{\textbf{0.27}}&
        \raisebox{1.5 em}{0.05}\\

        &{\includegraphics[width=15mm, height=15mm]{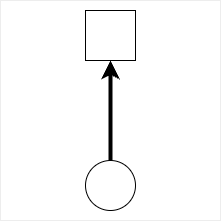}}&
        \raisebox{1.5 em}{-0.23}&
        \raisebox{1.5 em}{0.02}&
        \raisebox{1.5 em}{\textbf{0.11}}&
        \raisebox{1.5 em}{-0.03}\\
        
        &{\includegraphics[width=15mm, height=15mm]{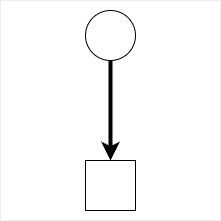}}&
        \raisebox{1.5 em}{0.49}&
        \raisebox{1.5 em}{0.30}&
        \raisebox{1.5 em}{\textbf{0.53}}&
        \raisebox{1.5 em}{0.44}\\

\bottomrule
\end{tabular}}
\caption{Spearman correlation between model answers and human answers.  \textbf{bold} = highest correlation}
\label{tab:visual}
\end{table*}
\renewcommand{\arraystretch}{1}

\end{document}